\documentclass[runningheads]{llncs}
\usepackage[T1]{fontenc}
\usepackage{graphicx,verbatim}
\usepackage{booktabs}
\usepackage{diagbox}
\usepackage{amsmath}
\usepackage{amssymb}
\usepackage{wrapfig}
\usepackage{multirow}
\usepackage{pifont}
\usepackage{wrapfig}
\usepackage[table]{xcolor}

\newcommand{\xmark}{\ding{55}}
\begin{document}
\title{Learning Cardiac Electrophysiology Digital Twins Through Agentic Discovery of Hybrid Structure}
%
\author{Ziqi Zhou\inst{1} \and
Yubo Ye\inst{1} \and
Sumeet Atul Vadhavka\inst{1}
\and \\ Linwei Wang\inst{1} \and Zhiqiang Tao\inst{1}
}

\authorrunning{Z. Zhou et al.}
%
\institute{Rochester Institute of Technology, Rochester, NY, 14623, USA
\\
\email{\{ez3783,yy8339,sv6234,lxwast,zxtics\}@rit.edu}
}


  
\maketitle              
\begin{abstract}

Building personalized cardiac electrophysiology (EP) digital twins requires identifying the appropriate model structure for each patient, not merely fitting parameters. Traditional methods rely on experts to manually prescribe hybrid physics-neural architectures, which requires deep domain expertise and does not transfer across patients. Recent works have applied large language models (LLMs) to generate or act as hybrid models. However, despite their promising generalization capacity,  these LLM-based methods lack the structural priors needed for stable cardiac simulations. Hence, we propose LEADS, a framework that formulates cardiac EP domain knowledge as a structured action space and utilizes an LLM agent to discover hybrid models. The agent follows an iterative \emph{reasoning-and-action} loop to select, combine, and refine hybrid models, whilst gradient descent handles parameter fitting. The proposed LEADS designs every candidate model towards physically grounded, interpretable, and numerically stable, while allowing open-ended architectural discovery. We validate LEADS on synthetic data with three ground-truth reaction models and on real cardiac EP data, demonstrating that it outperforms both human-designed hybrid models and other LLM-based hybrid modeling.


\end{abstract}
\section{Introduction}

Cardiac electrophysiology (EP) digital twins are personalized computational models that simulate a patient's cardiac electrical activity. They have broad clinical value: guiding catheter ablation for arrhythmias~\cite{prakosa2018personalized}, assessing sudden cardiac death risk~\cite{arevalo2016arrhythmia}, and supporting in silico drug testing~\cite{passini2017human}. However, building accurate EP digital twins remains challenging. Cardiac electrical propagation is typically modeled as a reaction-diffusion system~\cite{clayton2011models} on a discrete heart mesh, where the diffusion component describes how activation wavefronts spread across tissue, and the reaction component captures local ionic kinetics. Both components vary across patients due to differences in tissue properties, fibrosis, and disease state. A model finetuned for one patient can easily fail on another due to personalized structural assumptions. This means we need new methods that can automatically discover the appropriate model structure — beyond merely fitting parameters — for each individual patient.

Current approaches generally fall into two paradigms. The first is \emph{human-designed}: experts first pick a reaction model (\emph{e.g.}, Aliev-Panfilov~\cite{aliev1996simple}, Rogers-McCulloch~\cite{rogers2002collocation}) and a diffusion operator, and then fit its parameters to data. This human-design approach is interpretable but labor-intensive, requiring deep domain expertise, and the fixed architecture cannot transfer across patients. Recent hybrid methods~\cite{yin2021augmenting,jiang2024hyper,toloubidokhti2025meta} augment physics with neural components to improve flexibility, but still rely on manually prescribed structures---the expert must decide what to make neural networks and what to keep physical. The second paradigm uses Large Language Models (LLMs) for automation. HDTwinGen~\cite{holt2024automatically} innovatively asks LLMs to generate arbitrary hybrid model code via evolutionary reasoning. However, its action space remains unconstrained on cardiac EP, failing to produce cardiac activation on either synthetic or real clinical data. The underlying reason is that cardiac EP requires strong structural priors (e.g., stable ODE integration, physically meaningful state variables) that unconstrained code generation cannot guarantee. On the other hand, using LLMs directly as a digital twin~\cite{amad2025continuously} avoids the modeling problem but sacrifices physical interpretability and mathematical transparency, which are essential for clinical explanations. These limitations point to a clear gap: the automatic discovery of patient-specific architectures from prior biophysical knowledge remains underexplored. 

In this study, we propose \textbf{LEADS} (\textbf{L}earning \textbf{E}lectrophysiology hybrid model through \textbf{A}gentic \textbf{D}iscovery of \textbf{S}tructure). The key idea is to introduce cardiac EP modeling knowledge into a \emph{structured action space}, enabling an LLM agent to search over this space and thereby build the personalized hybrid model for each patient. LEADS bridges the two paradigms above: the structured space provides the domain priors that unconstrained LLM generation lacks, while the agent automates the architecture search that human experts do manually. Inspired by reasoning-and-act agent~\cite{yao2022react}, LEADS works in an iterative loop: it observes how previous candidate hybrid models performed, reasoning about what to improve (\emph{e.g.}, overfitting or insufficient capacity), and takes an action to produce an improved model (\emph{e.g.}, swapping the diffusion module, switching the reaction model, or rewriting code). For each candidate the agent proposes, gradient descent fits the model parameters on cardiac data. The structured catalog greatly reduces the chance of generating invalid models, while the agent retains the freedom to create novel hybrid structures beyond the predefined spaces. In summary, our work makes the following contributions:
\begin{itemize}
    \item We proposed a new LLM agent-based solution for the clinically important problem of automatically constructing personalized cardiac EP digital twins, mitigating the need for manual architecture design by domain experts.
    \item We introduce LEADS, which combines LLM agent reasoning and search with a structured hybrid action space that encodes domain priors as composable building blocks, achieving both search efficiency and physical interpretability.
    \item We validate LEADS on both synthetic data with three ground-truth reaction models and real clinical EGM recordings from the Utah dataset, demonstrating that LEADS outperforms human-designed hybrid models without requiring oracle knowledge of the underlying cardiac dynamics.
\end{itemize}
\section{Background}

\paragraph{\textbf{Hybrid Digital Twins.}}
Purely physics-based models of cardiac EP rely on well-known reaction-diffusion equations but assume fixed functional forms that may not match a given patient. Purely data-driven models can fit complex patterns but need large datasets and lack interpretability. Hybrid approaches try to get the best of both: they keep the physics where it is reliable and let a neural network handle the rest. APHYNITY~\cite{yin2021augmenting} introduced this idea for general dynamical systems by decomposing dynamics into a known physical term plus a learned residual. Follow-up work has improved robustness through synthetic data augmentation~\cite{wehenkel2022robust} and latent-space integration of expert ODEs~\cite{qian2021integrating}. In cardiac EP, HyPer-EP~\cite{jiang2024hyper} applies this decomposition with a meta-learning scheme to generalize across different tissue conditions, and~\cite{toloubidokhti2025meta} further extends it for cross-patient adaptation. A common limitation across all these methods is that the hybrid structure itself---which component is physical, which is neural, and how they interact---must be manually designed by a domain expert before training begins.

\paragraph{\textbf{LLM-Based Automated Discovery.}}
An alternative direction is to let machines design the model automatically. HDTwinGen~\cite{holt2024automatically} pioneered this for digital twins: it uses an LLM to propose, code, and iteratively refine model architectures through an evolutionary loop, with an evaluation agent providing feedback. This approach has shown strong results on simpler dynamical systems like pendulums. However, when applied to cardiac EP, the unconstrained code generation space becomes a liability---generated models often violate ODE stability requirements or fail to produce physiologically plausible outputs. Separately, some recent work uses LLMs not to build models but to \emph{be} the model: CALM-DT~\cite{amad2025continuously} frames digital twinning as in-context learning, and other efforts explore LLM-based simulation of human behavior~\cite{li2025far,du2025twinvoice} or autonomous task execution~\cite{li2023chattwin,shen2025mind}. While these demonstrate the breadth of LLM capabilities, they trade away the explicit mathematical structure that is needed for trustworthy clinical applications. These observations motivate combining the structural search capability of LLMs with domain-specific constraints to ensure physical validity.

\section{Methodology}

\paragraph{\textbf{Problem Definition.}}
\label{sec:problem}
Cardiac electrophysiology (EP) dynamics on a discrete heart mesh $\mathcal{G} = (\mathcal{V}, \mathcal{E})$ follow a reaction-diffusion ODE:
\begin{equation}
\frac{du}{dt} = \mathcal{D}(u, \mathcal{G}) + \mathcal{R}_u(u, v), \quad
\frac{dv}{dt} = \mathcal{R}_v(u, v),
\label{eq:reaction_diffusion}
\end{equation}
where $u, v \in \mathbb{R}^N$ are the transmembrane potential and recovery variable, respectively. $\mathcal{D}$ governs diffusion signal propagation, and $\mathcal{R}$ denotes local ionic reaction kinetics. Because exact forms of $\mathcal{D}$ and $\mathcal{R}$ vary across patients, classical approaches~\cite{aliev1996simple,rogers2002collocation,mitchell2003two} with fixed functional forms limit expressiveness. We aim to automatically discover both the optimal governing structures and parameters from observed trajectories.

\begin{figure}
    \centering
    \includegraphics[width=0.99\linewidth]{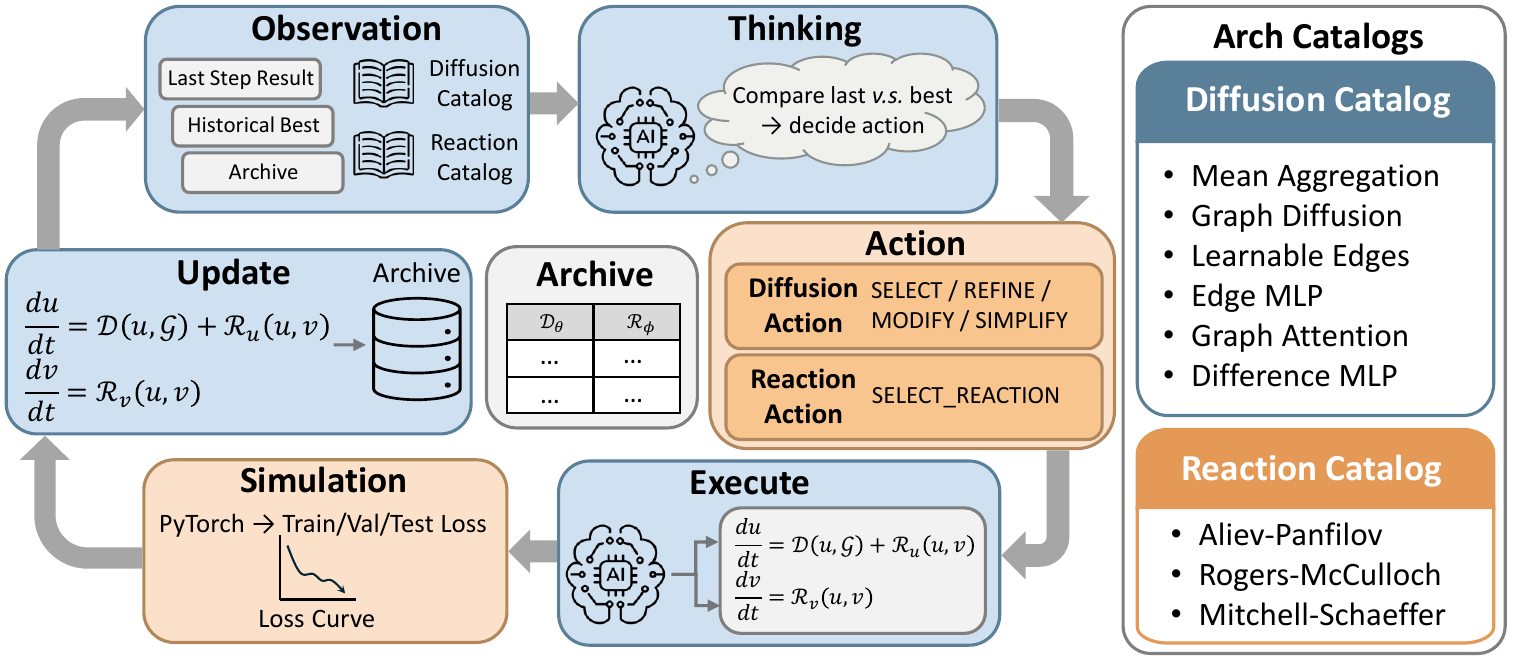}
    \caption{Overview of LEADS. Left: the LLM agent follows an Observe-Think-Act loop to iteratively assemble and refine hybrid cardiac EP models. Right: the structured action space with a neural diffusion catalog and a physics-based reaction catalog.}
    \label{fig:framework}
\end{figure}

\paragraph{\textbf{LEADS Overview.}}
\label{sec:leads_overview}
Cardiac EP is a domain rich in prior knowledge. The reaction-diffusion decomposition is physically well-established, and multiple reaction models have been proposed over decades of research. To leverage these domain priors rather than discarding them for black-box neural networks, we introduce \textbf{LEADS} (\textbf{L}earning \textbf{E}lectrophysiology hybrid model through \textbf{A}gentic \textbf{D}iscovery of \textbf{S}tructure).

As illustrated in Fig.~\ref{fig:framework}, LEADS uses an LLM agent to iteratively reason and search over a structured space of hybrid model components. The agent selects from a curated catalog of neural diffusion architectures and physics-based reaction models, and refines its choices based on empirical training feedback.

\paragraph{\textbf{Structured Action Space.}}
\label{sec:action_space}

Instead of generating arbitrary code like previous work~\cite{holt2024automatically}, LEADS navigates domain-informed architecture space. As shown in the right part of Fig.~\ref{fig:framework}, the space contains two base catalogs: a \textbf{Diffusion Catalog} (ranging from parameter-free mean aggregation to expressive graph attention networks) and a \textbf{Reaction Catalog} (established ionic models such as Aliev-Panfilov~\cite{aliev1996simple}, Rogers-McCulloch~\cite{rogers2002collocation}, and Mitchell-Schaeffer~\cite{mitchell2003two}).

At each step, the LLM agent makes two \emph{decoupled} decisions: selecting a reaction model from the catalog, and choosing one of four operations on the diffusion module:
\begin{itemize}
    \item \textsc{Select}: choose a new architecture from the diffusion catalog to explore an untried region of the design space.
    \item \textsc{Refine}: adjust hyperparameters (\emph{e.g.}, hidden dimensions, number of layers, activations) of an existing diffusion architecture.
    \item \textsc{Modify}: freely edit the underlying code of an existing diffusion module to create structural variants beyond the catalog templates, enabling open-ended architectural discovery.
    \item \textsc{Simplify}: reduce model complexity (\emph{e.g.}, removing layers or attention heads) to mitigate diagnosed overfitting.
\end{itemize}

Since reaction models are physics-based with fixed functional forms, only selection is needed; their learnable parameters are optimized by gradient descent (Sec.~\ref{sec:optimization}). This decoupled design allows the agent to independently explore each component---for example, systematically testing all reaction models against a fixed diffusion architecture, or refining the diffusion code while keeping a well-performing reaction model.

\begin{figure}[t]
    \centering
    \includegraphics[width=0.9\linewidth]{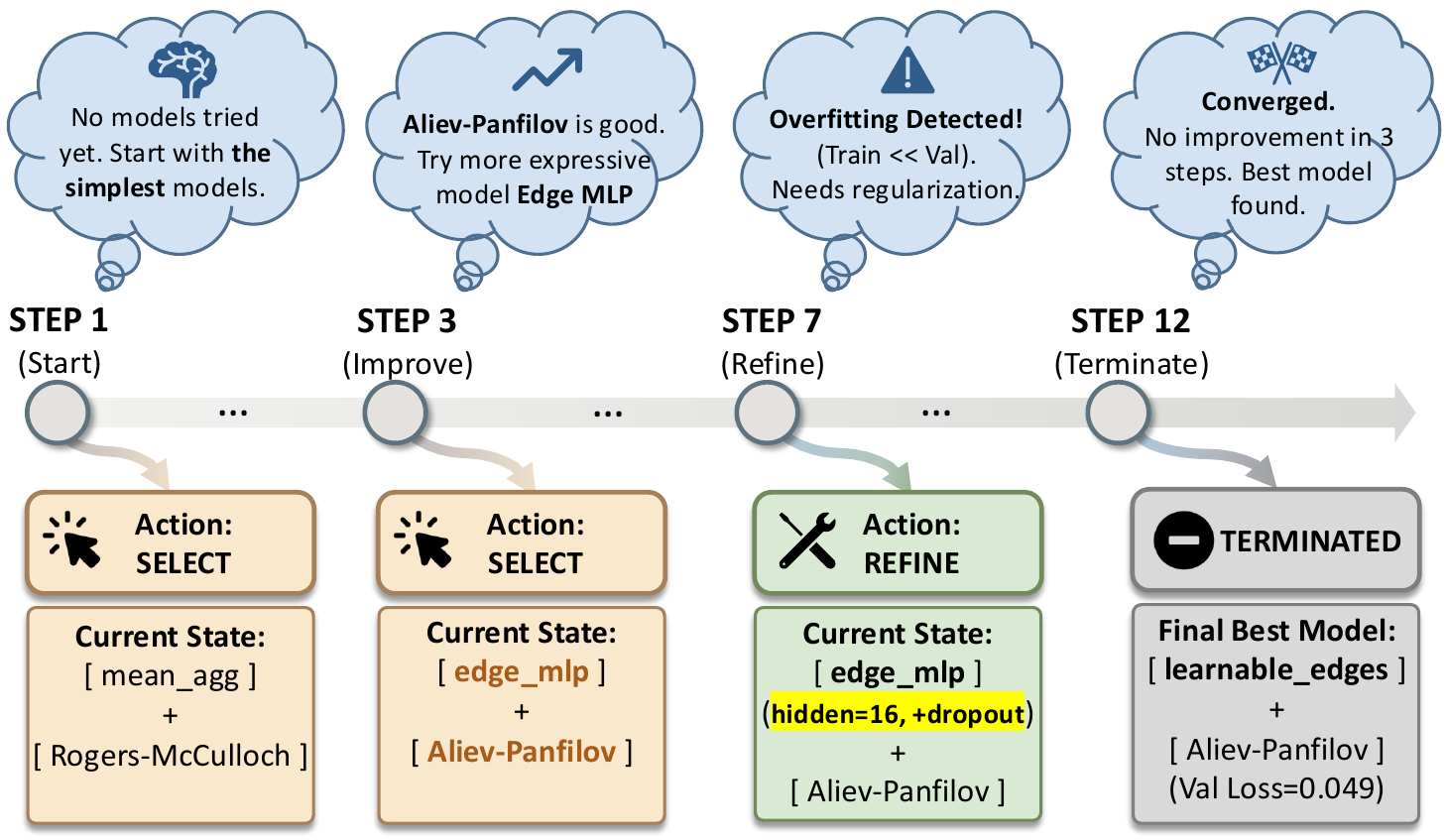}
    \caption{Example LEADS agent trajectory on AP synthetic data. The agent starts with simple models, switches to more expressive architectures, diagnoses overfitting and applies regularization, and terminates after convergence.}
    \label{fig:agent_loop}
\end{figure}

\paragraph{\textbf{Agentic Model Discovery.}}
\label{sec:optimization}
The design space of hybrid cardiac EP models is combinatorial: any diffusion architecture can be paired with any reaction model. Exhaustive enumeration is infeasible, and random search wastes budget on poor candidates. LEADS addresses this by using the LLM agent as an informed searcher that learns from prior evaluations.

At each step, the agent executes an \textbf{Observe-Think-Act} cycle. It first \emph{observes} a maintained archive $\mathcal{M}$ containing the validation losses and per-epoch training curves of all previous candidates. It then \emph{thinks} by generating an explicit reasoning trace---for instance, diagnosing overfitting, or identifying model capacity limits. Finally, it \emph{acts} by selecting one reaction model and applying one of the four diffusion operations (Fig.~\ref{fig:framework}) to produce a new hybrid structure. Fig.~\ref{fig:agent_loop} shows a concrete trajectory, the agent starts with simple models, progressively upgrades to more expressive and accurate hybrid models.

In each step of the LEADS loop, the candidate hybrid model parameters $(\theta, \phi)$ are fitted via gradient descent by minimizing the sum of squared trajectory errors:
$\mathcal{L}(\theta, \phi) = \sum_{k=1}^{T} \left\| \hat{u}(t_k; \theta, \phi) - u(t_k) \right\|^2.$
The agent handles structural decisions; gradient descent handles parameter fitting. The process terminates upon exhausting the action budget or converging, and returns the top-performing model from $\mathcal{M}$.

\section{Experiments}
\label{sec:experiments}

\subsection{Experimental Setup}
\label{sec:setup}

We evaluate LEADS on two cardiac EP settings: (1) Synthetic TMP trajectories generated from three ground-truth reaction models (AP~\cite{aliev1996simple}, RM~\cite{rogers2002collocation}, MS~\cite{mitchell2003two}) with Laplacian diffusion (32 samples each; 20/6/6 train/val/test split), and (2) Real EGM recordings from the Utah dataset~\cite{bergquist2021electrocardiographic} (20 samples; 12/3/5 split). We compare against two human-designed baselines with oracle access to the true synthetic reaction model---\textbf{Physics-Based} (Laplacian diffusion) and human-designed \textbf{Hybrid-Model}~\cite{jiang2024hyper} (Graph Attention Network~\cite{velivckovic2017graph})---as well as \textbf{HDTwinGen}~\cite{holt2024automatically}, an LLM-driven unconstrained code generator. All methods share identical mesh geometries (1119 nodes), data splits, and optimization settings. We utilize \textit{Gemini-2.5-Flash}~\cite{comanici2025gemini} as the LLM agent both in HDTwinGen~\cite{holt2024automatically} and LEADS experiments. Specifically, LEADS executes up to 15 ReAct steps with 3 inner-loop epochs per candidate, fine-tuning the best model for 10 epochs after the agentic loop. To prevent prior knowledge exploitation, reaction model names are anonymized in the catalog during the loop.

\subsection{Synthetic Data Experiments}
\label{sec:synthetic}

\begin{table}[t]
    \centering
    \begin{minipage}[b]{0.49\textwidth}
        \centering
        \caption{Test MSE ($\times 10^{-3}$, $\downarrow$) on synthetic data. $^*$ indicates the \textit{upper-bound}.}
        \label{tab:synthetic}
        \setlength{\tabcolsep}{4pt}
        \resizebox{\linewidth}{!}{%
        \begin{tabular}{l|ccc|c}
        \toprule
        \textbf{Method} & \textbf{AP} & \textbf{RM} & \textbf{MS} & \textbf{Avg.} \\
        \midrule
        Physics-Based$^*$ & 0.008  & 0.114  & 0.001  & 0.041 \\
        Hybrid-Model~\cite{jiang2024hyper} & 55.3  & 26.6 & 30.9  & 37.6 \\
        HDTwinGen~\cite{holt2024automatically} & 359.4 & 400.9 & 488.5 & 416.3 \\
        \rowcolor{gray!20}
        LEADS (Ours) & \textbf{49.2}  & \textbf{10.1}  & \textbf{20.8}  & \textbf{26.7} \\
        \bottomrule
        \end{tabular}
        }
    \end{minipage}
    \hfill
    \begin{minipage}[b]{0.49\textwidth}
        \centering
        \caption{Test mean activation time error ($\downarrow$) on real data.
        }
        \label{tab:real}
        \setlength{\tabcolsep}{6pt}
        \resizebox{0.88\linewidth}{!}{%
        \begin{tabular}{l|ccc}
        \toprule
        \textbf{Method} & \textbf{AP} & \textbf{RM} & \textbf{MS} \\
        \midrule
        Physics-Based & 3.42 & 13.08 & 4.37  \\
        Hybrid-Model~\cite{jiang2024hyper} & 5.64 & 6.62 & 7.48 \\
        HDTwinGen~\cite{holt2024automatically}  & \multicolumn{3}{c}{\xmark}\\
        \rowcolor{gray!20}
        LEADS (Ours) & \multicolumn{3}{c}{\textbf{5.52}} \\
        \bottomrule
        \end{tabular}
        }
    \end{minipage}
\end{table}

\begin{figure}[t]
\centering
\includegraphics[width=0.85\linewidth]{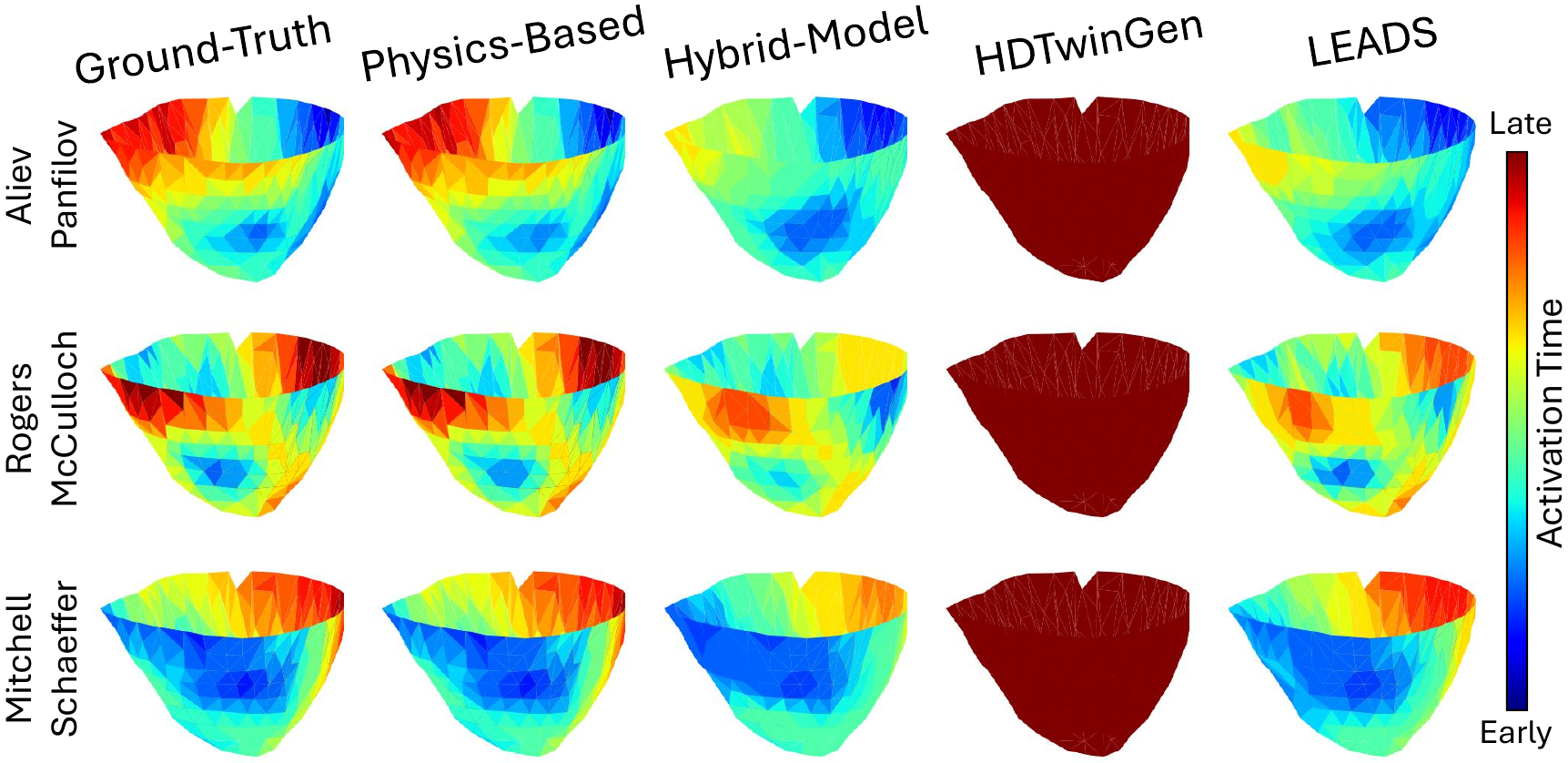}
\caption{Activation time (AT) maps on synthetic data. We show examples on AP, RM and MS generated data respectively. Colors encode AT from early (blue) to late (red). LEADS closely recovers the ground-truth patterns without true reaction knowledge.}
\label{fig:syn_vis_AT}
\end{figure}

Table.~\ref{tab:synthetic} and Fig.~\ref{fig:syn_vis_AT} summarize the quantitative and qualitative results on synthetic data. While the oracle Physics-Based model achieves near-zero test MSE ($\times 10^{-3}$), the human-designed Hybrid-Model struggles (avg.\ 37.6), highlighting the difficulty of learning a flexible diffusion operator. LEADS achieves a superior average MSE of 26.7. Despite lacking prior knowledge, LEADS correctly identifies the ground-truth reaction model for all three datasets (AP, RM, and MS) and optimizes the neural diffusion component to accurately reproduce ground-truth activation time (AT) wavefront patterns. In contrast, HDTwinGen (avg.\ MSE 416.3) fails to produce any cardiac activation, resulting in uniform AT maps. While HDTwinGen has demonstrated strong performance on simpler dynamical systems~\cite{holt2024automatically}, cardiac EP poses unique challenges---stable ODE integration on meshes, multi-variable coupling, and spatiotemporal wave propagation---that require domain-specific structural priors beyond what unconstrained code generation can provide.

\subsection{Real Data Experiments}
\label{sec:real}

\begin{figure}[t]
\centering
\includegraphics[width=0.85\linewidth]{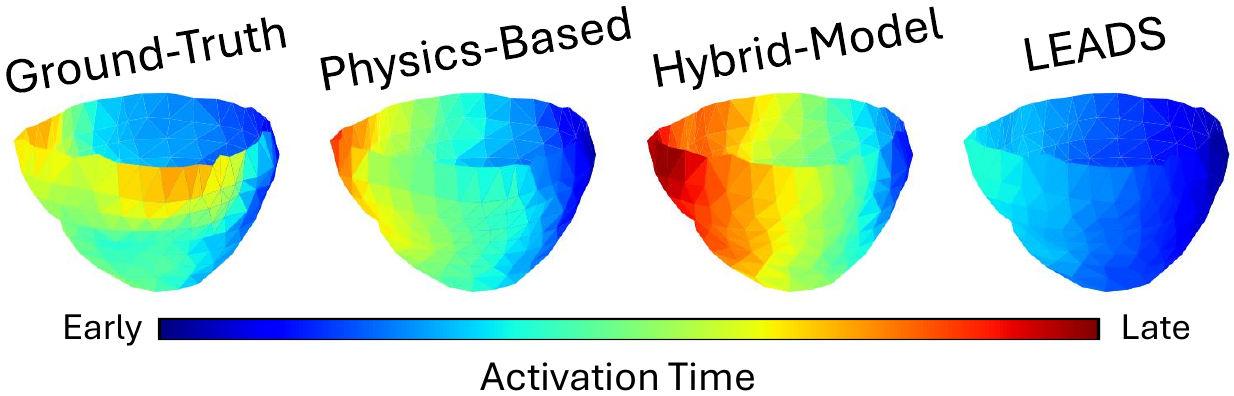}
\caption{Activation time maps on real data. Each column shows a different method. Colors encode AT from early (blue) to late (red). LEADS can produce activation patterns closer to the ground truth than human-designed hybrid model.}
\label{fig:utah_vis_AT}
\end{figure}

We further validate LEADS on the real dataset Utah~\cite{bergquist2021electrocardiographic}, which comes from animal model experiment. Lacking ground-truth kinetics and direct TMP observations, models are trained via a projected EGM loss and evaluated using test activation time (AT) MAE. As shown in Table.~\ref{tab:real}, while an optimally selected Physics-Based model (AP reaction) achieves the lowest AT error (3.42), the autonomous LEADS attains a competitive 5.52, outperforming the best human-designed Hybrid-Model (5.64). Crucially, visual AT maps in Fig.~\ref{fig:utah_vis_AT} reveal that the optimally selected Physics-Based model most accurately predicts activation pattern compared to the ground truth AT maps. In contrast, the human-designed Hybrid-Model exhibits a pronounced activation time delay. Meanwhile, LEADS produces smooth activation maps that align more closely with the ground-truth patterns than human-designed Hybrid-Model.

\subsection{Ablation Analysis}
\label{sec:ablation}
We ablate key design choices of LEADS on all three synthetic datasets (Table.~\ref{tab:ablation_syn}).
Fixing the reaction to the ground-truth (\textit{Diff-Only}) performs
\begin{wraptable}{r}{0.55\textwidth}
\vspace{-15pt}
\centering
\caption{Ablation on structured action space (test MSE, $\times 10^{-3}$, $\downarrow$).}
\label{tab:ablation_syn}
\setlength{\tabcolsep}{6pt}
\begin{tabular}{l|ccc}
\toprule
\textbf{Variant} & AP & RM & MS \\
\midrule
Diff-Only & 50.4  & 10.3 & 18.5 \\
React-Only & 49.6 & 20.4 & 20.7 \\
All-Neural & 66.1 & 25.5 & 25.9 \\
LEADS & \textbf{49.2}  & \textbf{10.1}  & \textbf{20.8} \\
\bottomrule
\end{tabular}
\vspace{-12pt}
\end{wraptable}
comparably on RM and MS but degrades on AP, showing oracle kinetics still require expressive diffusion. Conversely, fixing diffusion to Graph Attention (\textit{React-Only}) worsens RM error (20.4 vs.\ 10.1), showing that a fixed diffusion architecture bottlenecks performance even when the true reaction model is available. Furthermore, replacing the physics catalog with a fully data-driven module (\textit{All-Neural}) consistently degrades performance, particularly on AP (66.1 vs.\ 49.2) and RM (25.5 vs.\ 10.1), demonstrating that physics-based reaction priors are indispensable under limited training data. Overall, the full LEADS system achieves optimal performance across all datasets, confirming that \emph{joint} search over a \emph{hybrid} physics-neural catalog is essential for robust cardiac EP modeling.

\section{Conclusion}

We presented LEADS, an agentic framework that automates the discovery of hybrid cardiac EP digital twins. By providing an LLM agent with a domain-grounded catalog of diffusion and reaction components, LEADS turns hybrid model design into an iterative search problem guided by the agent's reasoning. This avoids both the labor of manual architecture design and the failure modes of unconstrained LLM-based methods. On synthetic data with three ground-truth reaction models, LEADS outperforms human-designed hybrid baselines despite having no oracle reaction knowledge. On real clinical electrogram recordings, LEADS achieves competitive activation time accuracy, surpassing the best human-designed hybrid model. Our current evaluation uses a single heart mesh geometry and a fixed catalog. Extending LEADS to multi-geometry settings, enriching the catalog with more diverse components, and applying it to other physiological domains are promising future directions.

%
%
%
\bibliographystyle{splncs04}
\bibliography{mybibliography}

\end{document}